\documentclass[10pt,twocolumn,letterpaper]{article}

\usepackage{iccv}
\usepackage{times}
\usepackage{epsfig}
\usepackage{graphicx}
\usepackage{amsmath}
\usepackage{amssymb}
\usepackage{bbm}
\usepackage{subfigure}
\usepackage{booktabs} 
\usepackage{stfloats}
\usepackage{algorithmic}
\usepackage{algorithm}
\usepackage{multirow}
\usepackage[T1]{fontenc}
\usepackage[breaklinks=true,bookmarks=false]{hyperref}
\usepackage{hyperref}
\hypersetup{
	colorlinks=true,
	filecolor=red,      
	citecolor=green,
}
\iccvfinalcopy 

\def\iccvPaperID{****}

\ificcvfinal\fi
\begin{document}

\title{Late Stopping: Avoiding Confidently Learning from Mislabeled Examples}

\author{Suqin Yuan\textsuperscript{1} \quad Lei Feng\textsuperscript{2}\footnotemark[1] \quad Tongliang Liu\textsuperscript{1}\footnotemark[1] \\
\textsuperscript{1}The University of Sydney \quad
\textsuperscript{2}Nanyang Technological University\\
}

\maketitle

\ificcvfinal\thispagestyle{empty}\fi

\renewcommand{\thefootnote}{\fnsymbol{footnote}} 
\footnotetext[1]{Corresponding authors.} 

\begin{abstract}
   Sample selection is a prevalent method in learning with noisy labels, where small-loss data are typically considered as correctly labeled data. However, this method may not effectively identify clean hard examples with large losses, which are critical for achieving the model's close-to-optimal generalization performance.  
   In this paper, we propose a new framework, Late Stopping, which leverages the intrinsic robust learning ability of DNNs through a prolonged training process.
   Specifically, Late Stopping gradually shrinks the noisy dataset by removing high-probability mislabeled examples while retaining the majority of clean hard examples in the training set throughout the learning process. 
   We empirically observe that mislabeled and clean examples exhibit differences in the number of epochs required for them to be consistently and correctly classified, and thus high-probability mislabeled examples can be removed.
   Experimental results on benchmark-simulated and real-world noisy datasets demonstrate that the proposed method outperforms state-of-the-art counterparts.
   \vskip -0.15in
\end{abstract}

\section{Introduction}

Deep Neural Networks (DNNs) have achieved outstanding success in various tasks, while the success largely relies on data with high-quality annotations \cite{han2020survey, 9941371, song2022learning, huang2022harnessing}. In many real-world scenarios, it would be quite difficult to collect large-scale accurately labeled data, which could inevitably contain noisy labels. Unfortunately, previous studies \cite{arpit2017closer, zhang2021understanding} showed that DNNs can easily overfit random labels, resulting in poor generalization performance. Therefore, an increasing number of methods have been proposed for Learning with Noisy Labels (LNL).

One mainstream solution in existing methods of LNL is to train the classifier with confident examples \cite{jiang2018mentornet, han2018co, wei2020combating, nguyen2019self, pleiss2020identifying}, which is based on the \emph{memorization effect} of DNNs, i.e., DNNs learn example with dominant patterns first and then overfit rare ones \cite{arpit2017closer}. Given only noisy data, to exploit the memorization effect, the typical strategy starts from a small confident clean dataset and then gradually expands the dataset, which prevents DNNs from over-fitting noisy data. In general, there are two primary approaches to exploit confident examples in the learning process. The first approach involves identifying examples with high-probability clean labels and training the classifier based on these examples, which is commonly referred to as the ``small-loss trick'' \cite{jiang2018mentornet, han2018co, han2020sigua, liu2020peer, xia2021sample}. The second approach involves controlling the learning process of the classifier to primarily learn high-probability clean examples in noisy datasets, which is commonly referred to as ``early stopping'' \cite{rolnick2017deep, tanaka2018joint, hu2019simple, li2020gradient, bai2021understanding}.

Despite providing satisfactory performance, these approaches present an unintended consequence in their methods to prevent over-fitting noise. Specifically, to reduce the impact of mislabeled examples, these approaches limit the capability of the model to effectively learn \emph{clean hard examples} (CHEs) in noisy datasets, where CHEs are defined as clean examples that are close to the decision boundary, and a significant proportion of CHEs are non-dominated sub-population examples \cite{feldman2020does}.
Identifying CHEs in noisy data is quite challenging \cite{bai2021me, karim2022unicon}, as both the CHEs and the mislabeled examples are often characterized by large losses \cite{xia2021sample, hacohen2019power, cao2020heteroskedastic}, causing them to become entangled.
To maintain the purity of the dataset of confident examples, existing LNL methods \cite{nguyen2019self, bai2021me} normally try to eliminate potential examples that are likely to be mislabeled, which inevitably contain many CHEs. 
However, it is necessary to consider the positive impact of memorization effects on underrepresented sub-populations \cite{feldman2020neural} (i.e., CHEs) when learning from natural datasets, as this is crucial for achieving close-to-optimal generalization performance.

\begin{figure*}[t]
\vskip -0.0in
\centering
\includegraphics[scale=0.71]{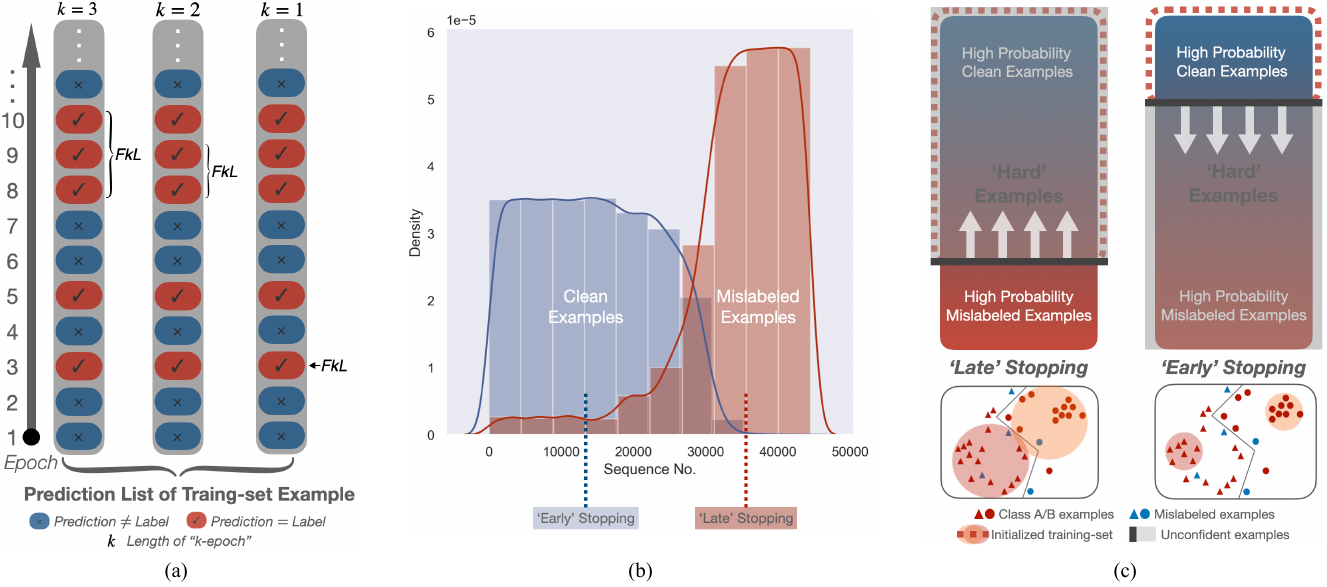}
\caption{
(a) We propose the \emph{First-time k-epoch Learning} (FkL) metric, which determines the minimum index of the training epoch until which an example has been predicted as its given label for consecutive $k$ epochs. 
(b) The normalised histogram of CIFAR-10 examples with 40\% symmetric label noise w.r.t. the sequence they meet the FkL metric during training procedure. The horizontal axis represents the sequential order in which training examples meet the FkL metric.  The vertical axis represents the normalised histogram of examples. 
(c) Rather than the methods that start from a small yet clean training set (‘Early’ Stopping), our proposed framework starts from a large training set (‘Late’ Stopping) that retains as many clean examples as possible.}
\label{fig1}
\vskip -0.10in
\end{figure*}

In this work, our goal is to enable the classifier to learn as many useful non-dominated sub-population examples in the training set as possible during the process of learning with noisy labels. This entails selecting as many clean examples as possible, particularly the clean hard examples, during the sample selection process. In relation to this context, we introduce a novel concept termed \emph{First-time k-epoch Learning} (FkL), which is defined as the index of the epoch during the training procedure where an example has been predicted to its given label for consecutive $k$ epochs for the first time, as shown in Figure \ref{fig1}(a).

\clearpage

Using \emph{First-time k-epoch Learning} (FkL) as a metric, we sequence the examples in the noisy dataset according to the order they meet the FkL metric during the training procedure, as shown in Figure \ref{fig1}(b). 
As shown in the same figure, most mislabeled examples can only be classified into their given labels for consecutive $k$ epochs in the later stages of training (i.e., larger in the sequence), which implies a relatively large FkL value.
This observation suggests that the examples with large FkL values (i.e., those examples that are classified to their given labels for consecutive $k$ epochs only in the late training stage) are predominantly those with incorrect labels. Therefore, the FkL values of different training examples can be used to distinguish whether an example is mislabeled or not.

Motivated by the above observations, we propose a novel method based on reverse thinking of the conventional confident example selection strategy. 
Our proposed method, called \emph{Late Stopping}, employs an iterative sample selection process that gradually reduces the noise rate of the dataset, leading to a positive feedback loop. 
Instead of selecting high-probability clean examples in the early stage, our method focuses on \emph{selecting high-probability mislabeled examples in the late stage} to retain as many CHEs as possible in the training set throughout the learning process, even though this method may lead to the retention of an acceptable level of mislabeled examples, as shown in \ref{fig1}(c).
Building on this, we introduce a novel sample selection criterion, FkL, which works effectively in selecting mislabeled examples under the \emph{Late Stopping} framework, and empirical results show that the FkL is more effective than the \emph{loss} criterion. 
We evaluate our method on benchmark-simulated and real-world noisy datasets. Empirical results demonstrate that the \emph{Late Stopping} method achieves superior performance compared with state-of-the-art counterparts of learning with noisy labels.

\section{Related Work}
\label{sec4}
Existing methods of learning with noisy labels can be broadly categorized into two types. The first category includes classifier-consistent algorithms that use the noise transition matrix \cite{liu2015classification, luo2017learning, xu2019l_dmi, patrini2017making, chen2019understanding}, which indicates the probabilities of clean labels flipping to noisy labels. The second category focuses on heuristic methods to mitigate the impact of label noise \cite{pleiss2020identifying, wu2021ngc, shen2019learning, wang2018iterative, lyu2019curriculum, wu2020topological, pleiss2020identifying}, which is also the primary focus of our work. Many existing heuristic methods in LNL are based on the memorization effect, where the algorithm attempts to only learn from confident clean examples \cite{jiang2018mentornet, han2018co, wei2020combating, nguyen2019self}. 
Alternatively optimizing the classifier and updating the training set is not a new idea in LNL. For instance, Joint Optim \cite{tanaka2018joint}, Co-teaching \cite{han2018co}, SELF \cite{nguyen2019self}, and Me-Momentum \cite{bai2021me} employ a similar positive feedback loop to our proposed \emph{Late Stopping}.

Previous research on the dynamics of DNNs has demonstrated that hard examples are typically learned during the late stage of the learning process \cite{arpit2017closer, toneva2018empirical, maini2022characterizing}. Furthermore, many recent studies have employed various training-time metrics to quantify the ``hardness'' of examples \cite{mangalam2019deep, hooker2019compressed, carlini2019distribution, baldock2021deep, jiang2021characterizing, maini2022characterizing}, leading to an increase in LNL approaches that use learning dynamics to select clean examples.
For instance, FSLT\&SSFT \cite{maini2022characterizing}, Self-Filtering \cite{wei2022self}, SELFIE \cite{song2019selfie}, and RoCL \cite{zhou2021robust} adopt a criterion similar to our proposed FkL criterion. Specifically, FSLT\&SSFT and Self-Filtering use the classifier's error prediction to pinpoint clean examples, SELFIE employs entropy values from prediction histories for selection, and RoCL leverages the variance of training losses to select clean examples. Instead of quantifying the ``hardness'' of examples based on the characteristics of individual examples in the dataset, our proposed method focuses on extracting the intrinsic robust learning ability of DNNs from the model trained on noisy datasets.

\section{Late Stopping}

\begin{algorithm*}[ht]
   \caption{Late Stopping}
   \label{alg1}
{\bfseries Input:} Original noisy training set $\mathcal{D}_1$, iteration rate $m\%$, noise rate $n\%$, epoch ${T_{max}}$ and iteration ${I_{max}}$.

{\bfseries Output:} Extracted final training set and the classifier.
\begin{algorithmic}[1]
   \FOR{${I}$ = 1, ..., ${I_{max}}$}
   \STATE {\bfseries Initialize} $\mathcal{D}_i$, $S_{F_{i}}$ = 0 and $\mathcal{D}_{F_{i}}$ = [ ].   
   \STATE//Initialize new training set, the number/dataset of examples that meet the \emph{First-time k-epoch Learning}, respectively.
   \FOR{${T}$ = 1, ..., ${T_{max}}$}
   \STATE {\bfseries Train} ${f_{i}}$ on $\mathcal{D}_i$.
   \STATE {\bfseries Update} $\mathcal{D}_{F_{i}}$, $S_{F_{i}}$.
   \STATE {\bfseries Break} and {\bfseries Output} $\mathcal{D}_{i+1}$ = $\mathcal{D}_{F_{i}}$, if $S_{F_i}$ $>$ $S_{F_{i-1}} \times (1-m\%)$.
   \ENDFOR
   \STATE {\bfseries Break} and {\bfseries Output} ${f_{i}}$ and $\mathcal{D}_{F_{i}}$, if $m \times i $ $>$ $n$.
   \ENDFOR
\end{algorithmic}
\end{algorithm*}

In this section, we elaborate on our proposed \emph{Late Stopping} method. As shown in Figure \ref{fig1}, the primary goal of the \emph{Late Stopping} framework is to maximize the retention of clean examples in the training set throughout the learning process, enabling our method to learn clean hard examples (CHEs) from a noisy training set.
Therefore, we halt the training process in the late training stage, leveraging \emph{First-time k-epoch Learning} (FkL) to exploit the results of the intrinsic robust learning ability of DNNs for sample selection. Iteratively performing this operation allows for a gradual reduction of noise while maximizing the retention of clean samples in the training dataset.

\subsection{Algorithm Flow}
For a comprehensive explanation of \emph{Late Stopping}, we begin by formally defining \emph{First-time k-epoch Learning} (FkL). In learning with noisy labels, let us consider a model ${f_i}$ trained on a noisy training set $\mathcal{D}_i$ composed of $n$ training examples $\{\mathbf{x}_j, y_j\}_{j=1}^n$ where $y_j$ denotes the given label (which may not be true) of $\mathbf{x}_j$. After $t$ training epochs, the predicted label $\hat y_j^t $ for the instance $\mathbf{x}_ j$ can be obtained. 
Let $\text{acc}_j^t = \mathbbm{1}_{\hat{y}_j^t=y_j}$  denote a binary variable indicating whether ${f_i}$ predicts the given label $y_j$ of the instance $\mathbf{x}_j$ at epoch $t$. 
The FkL for the instance  $\mathbf{x}_ j$ is defined as the the minimum index of the training epoch that the instance $\mathbf{x}_j$ has been predicted to its given label $y_j$ for $k$ consecutive epochs:
\vskip -0.15in
$$\mathrm{FkL}_{j}=\underset{t^{*}\in t}{\operatorname{argmin}}\left(\mathrm{acc}_j^{t^{*}} \wedge...\wedge \mathrm{acc}_j^{(t^{*}-k+1)} = 1\right).$$
\vskip -0.04in

If \( \mathrm{FkL}_j = t^* \), it implies that the classifier \( f_i \) ``learns'' the instance \( \mathbf{x}_j \) after \( t^* \) epochs. Notably, if \( \mathbf{x}_j \) is ``learned'' by the classifier early in the training, it will have a small FkL value, however, if it is ``learned''  in the late stage, it will have a large FkL value. Based on our FkL definition, we observed that the majority of examples with large FkL values are mislabeled, as shown in Figure \ref{fig1}(a). This observation provides a rationale for selecting examples with large FkL values, those learned in the late training stage, as high-probability mislabeled examples.

With the FkL selection criterion, we present the algorithmic framework of \emph{Late Stopping}. At the low level (Steps 4-8 of Algorithm \ref{alg1}), the sample selection strategy follows a ``from easy to hard'' curriculum learning procedure \cite{bengio2009curriculum,huang2019o2u}. During the $i$-th iteration, the classifier \( f_i \) is trained on a new training set $\mathcal{D}_i$ (Step 5), and newly identified FkL examples are added to \( \mathcal{D}_{F_i} \) (Step 6). If the size of \( \mathcal{D}_{F_i} \), represented as \( S_{F_i} \), exceeds a predefined threshold, the training in this iteration will halt, and the output becomes \( \mathcal{D}_{F_i} \) (Step 7). At the higher level (Steps 1-10 of Algorithm \ref{alg1}), our method learns through a ``from hard to easy'' positive feedback loop, distinguishing it from most existing LNL approaches. 
The algorithm starts with initializing \( \mathcal{D}_1 \) and training \( f_1 \) (Iteration $1$). In subsequent iterations, the new training set \( \mathcal{D}_i \) (Step 2, Iteration $i$) is updated from the previous iteration's \( \mathcal{D}_{F_{i-1}} \) (Step 8, Iteration \( i-1\)). When the stopping condition is met, the output is \( f_i \) and \( \mathcal{D}_{F_i} \) (Step 10).

\subsection{Learning CHEs by Positive Feedback Loop}
Here, we delve deeper into the typical positive feedback loop in LNL, highlighting its limitations in effectively learning CHEs. We then elucidate how our proposed method overcomes these limitations and maximizes the retention of clean examples throughout the learning process.

\textbf{Typical positive feedback loop in LNL.}
According to statistical learning theory \cite{mohri2018foundations}, training data with better quality yields a better classifier. Therefore, we can improve the classifier’s generalization performance in LNL through the construction of a positive feedback loop. The goal of the positive feedback loop is that, after each training step, the new classifier showcases enhanced generalization performance compared with its predecessor. By doing so, it can more robustly guide the sample selection process, enabling the use of improved training data for classifier training in the succeeding steps \cite{bai2021me}.

Most LNL methods, relying on confident examples, initiate with either a small yet clean training set or a weak yet reliable classifier. To enhance the classifier's generalization performance during initialization, they often rigorously limit memorization effects and eliminate potential mislabeled examples, even if this entails removing many CHEs. By restricting the memorization of mislabeled examples, they achieve superior generalization performance compared with unrestricted training on the original set. Then, they either iteratively or incrementally, train classifiers, ultimately obtaining either a larger clean training set or a more robust classifier. This strategy is effective when the classes or subclasses within the dataset are well-balanced and the examples within each class are interrelated, allowing the classifier to learn harder examples from easier ones.

\textbf{Limitations of typical positive feedback loop.}
Although LNL methods that rely on a typical positive feedback loop achieve good generalization performance in a variety of benchmark tasks, natural datasets are often highly imbalanced. This imbalance poses a challenge to achieving close-to-optimal generalization performance with existing LNL methods. Specifically, such challenges are because CHEs are often characterized by large losses and non-dominant patterns. As a result, substantial learning about CHEs predominantly transpires during the late stages of training, driven by memorization effects \cite{bai2021me, xia2021sample}. The conventional positive feedback loop, aimed at preventing overfitting, often strictly limits the memorization effects of DNNs. Consequently, this strict constraint reduces a classifier's ability to learn from under-represented subpopulations in datasets \cite{feldman2020does}, and thus limits its learning primarily to typical (dominant) classes and subclasses. Such constraints ultimately lead to a situation where the learning ability of rare classes and subclasses, i.e., CHEs, is continuously neglected, while the learning ability of the classes and subclasses representing the dominant pattern is continuously enhanced.

\begin{figure*}[t]
\vskip -0.1in
\begin{center}
\centerline{\includegraphics[width=16.5cm]{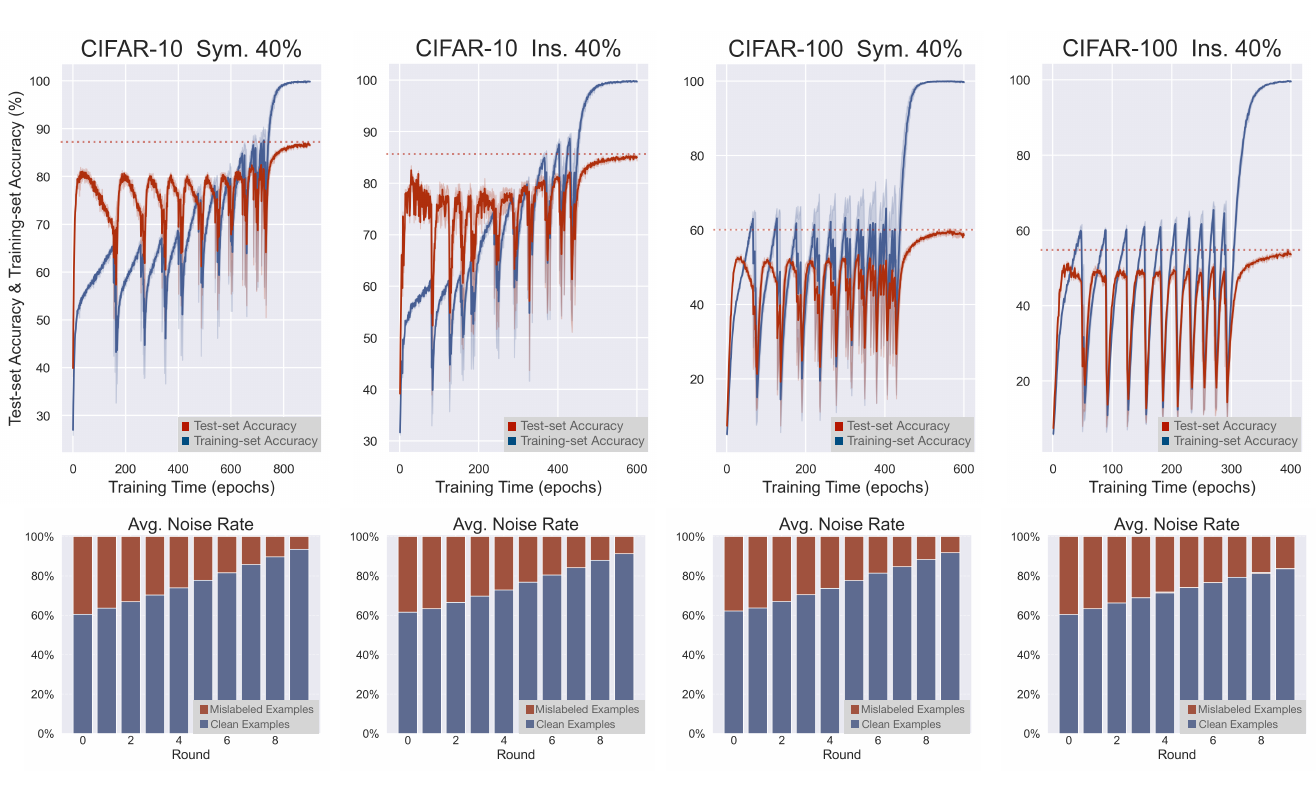}}
\caption{We refer to one update of the classifier and the training set as one iteration. Top subfigures: We illustrate the changes in the Test-set Accuracy and the Training-set Accuracy of the classifier during the training process of \emph{Late Stopping}. There are ten peaks in each figure because we set the number of iterations = 10 and the classifiers are re-initialized at the beginning of each new iteration. Bottom subfigures: We illustrate the changes in the noise rate of the training set in each iteration. There are ten bars in each figure because we set the number of iterations = 10 and the training set of each iteration is obtained from the previous iteration.}
\label{fig2}
\end{center}
\vskip -0.40in
\end{figure*}

\textbf{Positive feedback loop in \emph{Late Stopping}.}
Based on our previous discussions, to achieve close-to-optimal generalization performance on natural datasets, we need to focus on learning from CHEs throughout the training process. To achieve this, we prolong the training process for each training iteration to make use of the memorization effects to help memorize CHEs. With a loose sample selection process, we can ensure that we maximize the retention of clean examples in the training set throughout the learning process.
However, retaining as many clean examples as possible during the training initialization requires accepting a significant amount of label noise in the initial training set. Additionally, relying on the ``memorization effects in the late training stage'' to learn CHEs might result in poor generalization performance, as the classifier may seriously overfit label noise in this stage.

To counter these negative effects, we propose a ``from hard to easy'' positive feedback loop within the Late Stopping framework. While we prolong the training process in each training iteration, we simultaneously use the classifier's robust learning capabilities as the criterion for sample selection, i.e., FkL. In each iteration, we train a new classifier on a new training set, using it solely to guide the sample selection for the current iteration. In this positive feedback loop, we opt to disregard the generalization performance of the classifier in each iteration. Instead, we exploit the robust learning capabilities of each new classifier, \(f_i\), in its \(i\) iteration on \(\mathcal{D}_i\) from its learning dynamics, i.e., using the FkL criterion to select examples for the next iteration.

The new training set, obtained by the FkL criterion from the previous iteration, will have less noise. This enables the training of a better new classifier, which can learn more from the CHEs before it begins to memorize mislabeled examples. 
Therefore, tracking learning dynamics using the FkL criterion can better distinguish between CHEs and mislabeled examples, resulting in a new training set with even less noise than the previous one but contains CHEs. By observing each classifier's learning dynamics through FkL in every iteration, we can iteratively refine our training set, reducing noise while retaining the majority of clean examples, as shown in Figure \ref{fig2}.

\textbf{Discussion.}
Our proposed method is motivated by the seemingly conflicting views on the ability of DNNs to handle label noise as presented in recent studies: ``Deep neural networks easily fit random labels \cite{zhang2021understanding}'' and ``Deep learning is robust to massive label noises \cite{rolnick2017deep}''. 
Empirical evidence \cite{arpit2017closer} suggests that DNNs first learn simple patterns before over-fitting in the late training stage. However, there is a scarcity of studies on the behavior of DNNs during the intermediate stages of training, which may be a critical factor for understanding the robustness of DNNs to label noise - DNNs might exhibit spontaneously robust learning to massive label noises in all periods of training, though the generalization performance is over-fitting noisy labels.
Thus, we aim to separate the robust learning ability of DNNs from their generalization performance.
To this end, we leverage the intrinsic robust learning ability of DNNs by continuing the training process until the generalization performance seriously deteriorates (i.e., \emph{Late Stopping}) and we extract the robust learning results during this process (i.e., FkL).

\section{Experiments}
In this section, we conduct experiments on both synthetic and real-world datasets containing label noise to validate the effectiveness of our proposed \emph{Late Stopping}.

\textbf{Datasets.} 
We use two popular benchmark datasets i.e., CIFAR-10 and CIFAR-100 \cite{krizhevsky2009learning}, to test the generalization performance of our proposed method. Following previous works \cite{bai2021me,xia2020robust}, we manually corrupt CIFAR-10 and CIFAR-100 with Symmetric noise (abbreviated as Sym.) \cite{van2015learning} and Instance-dependent (abbreviated as Ins.) noise \cite{xia2020part}. For Sym. noise and Ins. noise, the noise rate is set to 20\% and 40\%. For a fair comparison, we leave out 10\% of the noisy training data.
We employ the real-world noisy dataset CIFAR-10N \cite{wei2021learning}, which equips the training datasets of CIFAR-10 with human-annotated real-world noisy labels.
More details about the above datasets and noise types are provided in Appendix A1.

\textbf{Compared methods.}  
We compare the \emph{Late Stopping} method (Algorithm \ref{alg1}) with the following well-known methods:
(1) Statistically inconsistent approaches (small-loss trick): MentorNet \cite{jiang2018mentornet}, Co-teaching \cite{han2018co}, and JoCoR \cite{wei2020combating};
(2) Statistically inconsistent approaches (other tricks): Joint Optim \cite{tanaka2018joint}, CDR \cite{xia2020robust}, and Me-Momentum \cite{bai2021me};
(3) Statistically consistent approaches: Forward \cite{patrini2017making} and DMI \cite{xu2019l_dmi}.
Note that drawing a direct comparison with certain semi-supervised approaches, especially those with multiple technical aggregations such as  SELF \cite{nguyen2019self}, DivideMix \cite{li2020dividemix}, ELR+ \cite{liu2020early}, and Unicon \cite{karim2022unicon}. The results for baselines are copied from original papers and \cite{bai2021me, wei2021learning}.

\textbf{Network structure and experimental setup.}
We utilize a typical warming-up strategy to ensure stable predictions of the DNNs.
The optimizer settings used in our experiments are as follows: SGD with a momentum of 0.9, weight decay of 5e-4, batch size of 128, and initial learning rate of 0.02. In our experiments, we applied typical data augmentations such as horizontal flipping and random cropping.
For the experiments on synthetic noisy datasets, ResNet-18 and ResNet-34 \cite{he2016deep} are used for CIFAR-10 and CIFAR-100, respectively.
For the experiments on real-world noisy datasets, ResNet-34 is used for CIFAR-10N.
More details about our experimental setup are provided in Appendix A2.

\subsection{Effectiveness of Late Stopping}

Here, we aim to provide empirical evidence to verify the effectiveness of \emph{Late Stopping}. We conduct experiments on the CIFAR-10 and CIFAR-100 datasets, as their ground-truth labels are available. 
The average results of five runs of our experiments are presented in Figure \ref{fig2}, which demonstrates the iterative improvement in the quality of the training set when \emph{Late Stopping} is applied in the learning process. And a classifier with better generalization performance is obtained.
We demonstrate the ability of our proposed FkL criterion to select mislabeled examples under the \emph{Late Stopping} framework, compared with the \emph{loss} criterion. 
Additionally, we show the effectiveness of our proposed \emph{Late Stopping} in retaining clean examples in the training set by comparing it with Me-Momentum \cite{bai2021me}, which is also an sample selection method that focuses on extracting clean hard examples.

\begin{table*}[t]
\centering

	\caption{Label precision of the selected examples (45$k$ in total, with 27$k$ clean examples).}
	\label{tab1}
\resizebox{1\textwidth}{!}{
\setlength{\tabcolsep}{4.7mm}{
\begin{tabular}{cc|ccc|ccc}
\toprule
& &\multicolumn{3}{c|}{Selecting Clean Examples} & \multicolumn{3}{c}{Selecting Mislabeled Examples}\\
\cmidrule(lr){3-5}\cmidrule(lr){6-8}
 Data\&Noise&criterion&0-10$k$ & 0-15$k$ & 0-25$k$ & 30$k$-45$k$ & 35$k$-45$k$ & 40$k$-45$k$ \\
\midrule
\multirow{1}{1.5cm}{CIFAR-10} & \emph{loss} & 92.73\% & 89.56\% & 82.89\% & 75.67\% & 83.49\% & 91.54\% \\
\emph{Sym. 40\%}& FkL(Ours) & \textbf{95.49\%} & \textbf{95.59\%} & \textbf{93.56\%} & \textbf{96.39\%} & \textbf{99.81\%} & \textbf{99.94\%} \\
\midrule
\multirow{1}{1.5cm}{CIFAR-10} & \emph{loss} & 75.55\% & 72.33\% & 71.38\% & 60.27\% & 68.69\% & 77.28\%\\
\emph{Ins. 40\%}& FkL(Ours) & \textbf{81.88\%} & \textbf{81.67\%} & \textbf{81.51\%} & \textbf{78.51\%} & \textbf{88.85\%} & \textbf{99.30\%} \\
\midrule
\multirow{1}{1.7cm}{CIFAR-100} & \emph{loss} & 88.43\% & 84.72\% & 78.49\% & 68.02\% & 75.73\% & 84.68\%\\
\emph{Sym. 40\%}& FkL(Ours) & \textbf{97.45\%} & \textbf{96.78\%} & \textbf{95.06\%} & \textbf{94.63\%} & \textbf{97.91\%} & \textbf{98.86\%} \\
\midrule
\multirow{1}{1.7cm}{CIFAR-100} & \emph{loss} & 71.30\% & 68.23\% & 65.29\% & 45.91\% & 44.23\% & 42.56\% \\
\emph{Ins. 40\%}& FkL(Ours) & \textbf{89.25\%} & \textbf{88.56\%} & \textbf{84.40\%} & \textbf{78.22\%} & \textbf{86.28\%} & \textbf{91.76\%} \\
\bottomrule  
\end{tabular}
}
}
\vskip -0.05in
\end{table*}

\begin{table*}[!t]
\centering

	\caption{Numbers of clean examples of the final training set (27$k$ in total).}
	\label{tab2}
\resizebox{1.0\textwidth}{!}{
\setlength{\tabcolsep}{1.8mm}{
\begin{tabular}{c|cc|cc}
\toprule
 & CIFAR-10 (\emph{Sym. 40\%}) & CIFAR-10 (\emph{Ins. 40\%}) & CIFAR-100 (\emph{Sym. 40\%}) & CIFAR-100 (\emph{Ins. 40\%}) \\
\midrule
Me-Momentum \cite{bai2021me} & 25,791 & 25,694 & 22,244 & 20,779 \\
Late Stopping (Ours) & \textbf{26,562} & \textbf{25,915} & \textbf{26,120} & \textbf{23,979} \\
\bottomrule  
\end{tabular}
}
}
\vskip -0.05in
\end{table*}

\textbf{Comparison with loss criterion.}
The selection of confident clean examples based on the \emph{small}-\emph{loss} criterion is a commonly used approach in LNL and has been shown to be effective \cite{jiang2018mentornet, han2018co, wei2020combating}. However, the approach of selecting mislabeled examples based on the \emph{large}-\emph{loss} \cite{cao2020heteroskedastic} can be tricky since it fails to distinguish between mislabeled and clean hard examples \cite{xia2021sample}.
We conducted experiments on the CIFAR-10 and CIFAR-100 datasets to compare the effectiveness of the FkL and \emph{loss} criterion for selecting examples during the same training process. Specifically, we ranked all examples (45$k$ in total, with 27$k$ clean examples) using the FkL criterion and the \emph{loss} criterion, respectively, from the smallest 0$k$ to the largest 45$k$. We then compared the precision of the two criteria in selecting clean examples and mislabeled examples in different ranges. The details of the experiments and the results are presented in Table \ref{tab1}.
Our experimental results demonstrate that the FkL criterion outperforms the \emph{loss} criterion in selecting both clean and mislabeled examples under the \emph{Late Stopping} framework. The largest performance gap was observed in the 40$k$-45$k$ range. These results provide empirical evidence that the FkL criterion is more effective than the \emph{loss} criterion for selecting mislabeled examples in the \emph{Late Stopping} framework.

\textbf{Comparison with hard example selection approach.}
In Table \ref{tab2}, we compare our proposed method with Me-Momentum \cite{bai2021me} which also focuses on selecting clean hard examples. Since there are no clear definitions of clean hard examples, we cannot directly compare the effectiveness of selecting such examples. Nevertheless, we can make an indirect comparison by comparing the ability to select clean examples.
Our experimental results demonstrate that \emph{Late Stopping} is an effective approach for retaining clean hard examples. On the CIFAR-10 dataset, both \emph{Late Stopping} and Me-Momentum showed comparable performance. However, on the more challenging CIFAR-100 dataset, \emph{Late Stopping} significantly achieved better performance in selecting clean hard examples.

\subsection{Classification Accuracy}
\label{sec42}
\textbf{Synthetic datasets.} 
The experimental results on test accuracy on synthetic datasets with class-dependent and instance-dependent label noise are provided in Table \ref{tab3} and \ref{tab4}. Each trial is repeated five times and the mean value and standard deviation are recorded. 
On CIFAR-10, our method achieves varying degrees of lead over baselines. In the 20\% symmetric noise task, which is the simplest task in all cases, the Me-Momentum baseline outperforms ours, which illustrates the effectiveness of conventional sample selection based on confident clean examples for simple LNL.
For the more challenging CIFAR-100 dataset, our proposed method consistently achieved the best results. The size of each class in CIFAR-100 is ten times smaller than that of CIFAR-10, making it difficult to retain sufficient CHEs while maintaining a small yet clean training set using conventional sample selection methods. Methods that focus on improving the number of CHEs in the training set can achieve better generalization performance in this task.
For instance, DMI and Co-teaching show a much larger performance gap with our proposed method on CIFAR-100 compared with CIFAR-10. Furthermore, Table \ref{tab2} validates our method's superior capability to retain clean examples, resulting in a larger performance gap with Me-Momentum on CIFAR-100.

\textbf{Real-world dataset.} 
To validate the efficacy of our proposed method on real-world datasets, we assess our method using the CIFAR-10N dataset in Table 4, a recognized benchmark in learning with noisy labels tasks with both ground-truth labels and human-annotated real-world noisy labels. We utilized the most challenging real-world noisy labels from CIFAR-10N, i.e., the ``\emph{Worst}'' setting. In this setting, for each image, if there are any incorrectly labeled examples, the given label is randomly selected from human-annotated false labels. Each trial is repeated five times and the mean value and standard deviation are recorded. As shown in Table \ref{tab4}, our method achieves the best performance compared with other methods.

\begin{table*}[!t]
\centering

	\caption{Test performance (mean$\pm$std) of each approach using ResNet-18 on CIFAR-10.}
	\label{tab3}
\resizebox{1\textwidth}{!}{
\setlength{\tabcolsep}{6.5mm}{
\begin{tabular}{ccccc}
\toprule

                      & Sym. 20\% & Sym. 40\% & Ins. 20\% & Ins. 40\%\\
                      \midrule
Late Stopping(Ours)  & 91.06$\pm$0.22\% & \textbf{88.92$\pm$0.38\%} & \textbf{91.08$\pm$0.23\%} & \textbf{87.41$\pm$0.38\%} \\
\midrule
Cross-Entropy \cite{rubinstein1999cross} & 85.00$\pm$0.43\% & 79.59$\pm$1.31\% & 85.92$\pm$1.09\% & 79.91$\pm$1.41\% \\
MentorNet \cite{jiang2018mentornet} & 80.49$\pm$0.11\% & 77.48$\pm$3.45\% & 79.12$\pm$0.42\% & 70.27$\pm$1.52\% \\
Forward \cite{patrini2017making} & 85.63$\pm$0.11\% & 74.30$\pm$0.26\% & 85.29$\pm$0.38\% & 74.72$\pm$3.24\% \\
Co-teaching \cite{han2018co} & 87.16$\pm$0.52\% & 83.59$\pm$0.28\% & 86.54$\pm$0.11\% & 80.98$\pm$0.39\% \\
JoCoR \cite{wei2020combating} & 88.69$\pm$0.19\% & 85.44$\pm$0.29\% & 87.31$\pm$0.27\% & 82.49$\pm$0.57\% \\
DMI \cite{xu2019l_dmi} & 88.18$\pm$0.13\% & 83.98$\pm$0.48\% & 89.14$\pm$0.36\% & 84.78$\pm$1.97\% \\
Joint Optim \cite{tanaka2018joint} & 89.70$\pm$0.36\% & 87.79$\pm$0.20\% & 89.69$\pm$0.42\% & 82.62$\pm$0.57\% \\
CDR \cite{xia2020robust} & 89.68$\pm$0.38\% & 86.13$\pm$0.44\% & 90.24$\pm$0.39\% & 83.07$\pm$1.33\% \\
Me-Momentum \cite{bai2021me} & \textbf{91.44$\pm$0.33\%} & 88.39$\pm$0.34\% & 90.86$\pm$0.21\% & 86.66$\pm$0.91\% \\
\bottomrule  
\end{tabular}
}
}
\end{table*}
\begin{table*}[!t]
\centering

	\caption{Test performance (mean$\pm$std) of each approach using ResNet-34 on CIFAR-100.}
	\label{tab4}
\resizebox{1\textwidth}{!}{
\setlength{\tabcolsep}{6.5mm}{
\begin{tabular}{ccccc}
\toprule

                      & Sym. 20\% & Sym. 40\% & Ins. 20\% & Ins. 40\%\\
                      \midrule
Late Stopping(Ours)  & \textbf{68.67$\pm$0.67\%} & \textbf{64.10$\pm$0.40\%} & \textbf{68.59$\pm$0.70\%} & \textbf{59.28$\pm$0.46\%} \\
\midrule
Cross-Entropy \cite{rubinstein1999cross} & 57.59$\pm$2.55\% & 45.74$\pm$2.61\% & 59.85$\pm$1.56\% & 43.74$\pm$1.54\% \\
MentorNet \cite{jiang2018mentornet} & 52.11$\pm$0.10\% & 35.12$\pm$1.13\% & 51.73$\pm$0.17\% & 40.90$\pm$0.45\% \\
Forward \cite{patrini2017making} & 57.75$\pm$0.37\% & 38.59$\pm$1.62\% & 58.76$\pm$0.66\% & 44.50$\pm$0.72\% \\
Co-teaching \cite{han2018co} & 59.28$\pm$0.47\% & 51.60$\pm$0.49\% & 57.24$\pm$0.69\% & 45.69$\pm$0.99\% \\
JoCoR \cite{wei2020combating} & 64.17$\pm$0.19\% & 55.97$\pm$0.46\% & 61.98$\pm$0.39\% & 50.59$\pm$0.71\% \\
DMI \cite{xu2019l_dmi} & 58.73$\pm$0.70\% & 49.81$\pm$1.22\% & 58.05$\pm$0.20\% & 47.36$\pm$0.68\% \\
Joint Optim \cite{tanaka2018joint} & 64.55$\pm$0.38\% & 57.97$\pm$0.67\% & 65.15$\pm$0.31\% & 55.57$\pm$0.41\% \\
CDR \cite{xia2020robust} & 66.52$\pm$0.24\% & 60.18$\pm$0.22\% & 67.06$\pm$0.50\% & 56.86$\pm$0.62\% \\
Me-Momentum \cite{bai2021me} & 68.03$\pm$0.53\% & 63.48$\pm$0.72\% & 68.11$\pm$0.57\% & 58.38$\pm$1.28\% \\

\bottomrule  
\end{tabular}
}
}

\end{table*}

\begin{table*}[!t]
\centering

	\caption{Test accuracy of each approach using ResNet-34 on CIFAR-10N (\emph{Worst}).}
	\label{tab5}
\resizebox{1.0\textwidth}{!}{
\setlength{\tabcolsep}{3mm}{
\begin{tabular}{cccccc}
\toprule
Cross-Entropy \cite{rubinstein1999cross} & Forward \cite{patrini2017making} & Co-teaching \cite{han2018co} & JoCoR \cite{wei2020combating} & Me-Momentum \cite{bai2021me} & Late Stopping(Ours)\\
\midrule
77.69$\pm$1.55\% & 79.79$\pm$0.46\% & 83.83$\pm$0.13\% & 83.37$\pm$0.30\% & 84.71$\pm$0.37\% & \textbf{85.24$\pm$0.38\%} \\
\bottomrule  
\end{tabular}
}
}

\end{table*}

\subsection{Further analysis}
Here, we conduct further analysis of our proposed method with the same experimental settings as those used in Section \ref{sec42} unless stated otherwise.

\begin{figure}[t]
\centering
\includegraphics[scale=0.73]{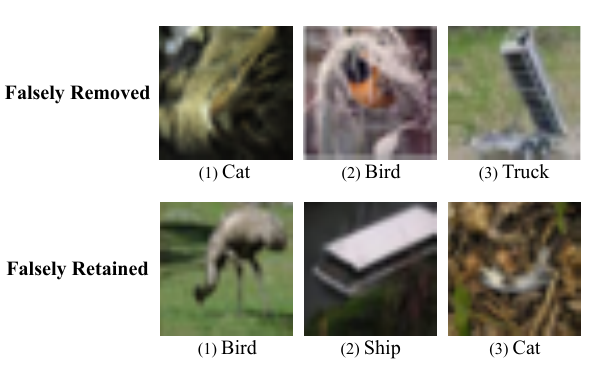}
\caption{The labels in the figure are ground-truth labels for each example. Falsely Removed (1-3) are typical examples that are incorrectly removed in the first iteration of Late Stopping; Falsely Retained (1-3) are typical examples that are incorrectly marked by FkL in the last iteration of Late Stopping. In the training set, Falsely Retained (1) is incorrectly labeled as ``deer'', Falsely Retained (2) is incorrectly labeled as ``truck'', and Falsely Retained (3) is incorrectly labeled as ``frog''.}
\label{fig3}

\end{figure}

\textbf{Limitations.} 
While our method achieves good robustness by simply removing the lately learned examples, it is not immune to errors. Some clean examples might be incorrectly removed from the training set, i.e., \emph{falsely removed examples}, and some mislabeled examples are incorrectly retained in the training set, i.e., \emph{falsely retained examples}. We visualize such typical examples in Figure \ref{fig3} and provide the following empirical analysis.

\emph{Falsely retained examples} are mislabeled with the wrong label $L_w$, however, such examples often possess patterns similar to the dominant subclasses in the $L_w$ class, making it difficult to identify them as high probability mislabeled examples by the FkL criterion. As shown in Table \ref{tab6}, based on the FkL criterion and the loss criterion, the ``hardness'' of these examples decreases after being mislabeled. In particular, we named the examples with a decrease in training losses after being mislabeled as ``\emph{hard mislabeled examples}''. Our experimental results indicate that \emph{hard mislabeled examples} pose a challenge for all sample selection-based LNL methods. More details about Table \ref{tab6} are provided in Appendix B1.

\emph{Falsely removed examples} refer to the examples that are too hard to remove in practice. To limit the computational resources consumed by our proposed method, we set the removal rate (Algorithm \ref{alg1}, iteration rate $m$\%) to a large value. Thus, our proposed method cannot completely distinguish between rare patterns and mislabeled examples, which can result in a small number of examples from rare subclasses being mistakenly removed from the training set.

\begin{table}[t]
\centering
	\caption{The comparison of the average ranking of \emph{falsely retained examples} using the \emph{loss} criterion and the FkL criterion before and after fixing the noisy labels (CIFAR-10, Before: \emph{Sym. 40\%} noise).}
	\label{tab6}
\resizebox{1\columnwidth}{!}{
\setlength{\tabcolsep}{3mm}{
\begin{tabular}{c|c|c}
\toprule

Label & Criterion & \emph{Avg.} Ranking\\
\midrule
Before fixing & FkL & 22340.66 \\
(Given label) & \emph{loss} & 23673.50 \\
 \midrule
After fixing & FkL & 28452.08 (\textbf{+27.36\%})\\
(Ground-truth label) & \emph{loss} & 29476.46 (\textbf{+24.51\%})\\
\bottomrule  
\end{tabular}
}
}
\end{table}
\begin{table}[t]
\centering
	\caption{Comparison of the number of clean examples in datasets before and after applying our method as a pre-processing approach to decrease the noise level of a 40\% noisy training set to 20\%.}
	\label{tab7}
\resizebox{1.0\columnwidth}{!}{
\setlength{\tabcolsep}{1.5mm}{
\begin{tabular}{c|cc}
\toprule
Training set & Before (\emph{40\% noise}) & After (\emph{20\% noise})\\
\midrule
CIFAR-10 (\emph{Sym.})& 27228 & 27139 (\textbf{-0.33\%}) \\
CIFAR-10 (\emph{Ins.}) & 27094 & 26710 (\textbf{-1.42\%}) \\
CIFAR-100 (\emph{Sym.})& 27341 & 27037 (\textbf{-1.11\%}) \\
CIFAR-100 (\emph{Ins.}) & 27249 & 25079 (\textbf{-7.96\%}) \\
CIFAR-10N (\emph{Worst}) & 29896 & 29613 (\textbf{-0.95\%}) \\
\bottomrule  
\end{tabular}
}
}

\end{table}

\begin{table}[t]
\centering
	\caption{Comparison of the total training hours on CIFAR-100.}
	\label{tab8}
\resizebox{1.0\columnwidth}{!}{
\setlength{\tabcolsep}{2mm}{
\begin{tabular}{cccc}
\toprule

 Co-teaching \cite{han2018co} & Me-Momentum \cite{bai2021me} & Late Stopping (Ours)\\
\midrule
3.3h & 7.1h & 9.5h \\
\bottomrule  
\end{tabular}
}
}
\vskip -0.15in
\end{table}

\textbf{Versatility.}
We are focusing on demonstrating the concept, i.e., using \emph{Late Stopping} to retain clean hard examples in the training set, but not on boosting the classification performance. However, to cope with even more complex noisy scenarios, our method works well in combination with other advanced techniques. In particular, our approach has excellent performance as a pre-processing approach for reducing the noise rate of the noisy training set. As shown in Table \ref{tab7}, we evaluate \emph{Late Stopping} as a pre-processing approach to decrease the noise level of a 40\% noisy training set to 20\%. Our experimental results indicate that our proposed method performs exceptionally well in reducing label noise. It can substantially reduce the noise level of the training set with minimal loss of clean examples.

\begin{figure}[t]
\centering
\includegraphics[scale=0.62]{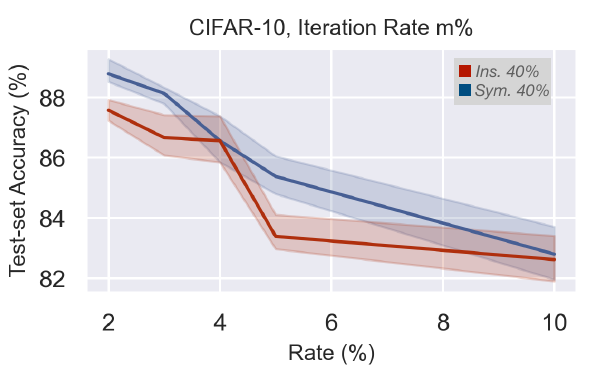}
\caption{Illustrates the performance of classifiers with the change in iteration rate $m$.}
\label{fig4}
\vskip -0.2in
\end{figure}

\textbf{Training time.} 
Our proposed method involves training the classifier on a larger training set for a longer period, which may not be advantageous in terms of efficiency. Despite this, as the noise rate decreases, the size of the training set decreases, and the speed of obtaining FkL-examples is accelerated (see Figure \ref{fig2}). As shown in Table \ref{tab8}, we compare the training time with typical baseline methods on CIFAR-100 (\emph{Sym. 40\%}). Our experimental results indicate that the training time of our proposed method is in a comparable range with other approaches.

\textbf{Sensitivity of the iteration rate.} The iteration rate $m$ (see Algorithm \ref{alg1}) determines the iteration number for \emph{Late Stopping} and the number of examples removed in each iteration. To investigate the sensitivity of $m$, we conducted experiments on CIFAR10 with 40\% Sym. noise and Ins. noise by varying $m$ in the range ${\{10, 5, 4, 3, 2\}}$. 
Figure \ref{fig4} shows the performance of classifiers, which gradually improves as $m$ decreases. This observation aligns with the findings presented in Table \ref{tab1}, where a decrease in the range of selected high-probability mislabeled examples leads to higher accuracy in selecting such examples, ultimately benefiting the classifier's performance.

\section{Conclusion}
In this paper, we focused on the challenge of the inability of existing sample selection methods to effectively select clean hard examples. To address this challenge, we proposed a novel method called \emph{Late Stopping}. In contrast to traditional early-stopping strategies, our method is rooted in a prolonged training process that distinguishes between mislabeled and clean examples by pinpointing the number of training epochs required for each example to be consistently classified to its given label for the first time. We coin this new sample selection criterion as \emph{First-time k-epoch Learning}. Experimental results on synthetic and real-world datasets demonstrate that our proposed method is both straightforward and effective in handling learning with noisy labels. In future work, we plan to enhance the FkL criterion to more accurately capture the intrinsic robust learning ability of DNNs, which could potentially bolster the effectiveness of our \emph{Late Stopping} framework, especially under more intricate noise conditions.

\section*{Acknowledgement}
Lei Feng was supported by the National Natural Science Foundation of China (Grant No. 62106028), Chongqing Overseas Chinese Entrepreneurship and Innovation Support Program, CAAI-Huawei MindSpore Open Fund, and Chongqing Artificial Intelligence Innovation Center. Tongliang Liu was partially supported by Australian Research Council Projects IC-190100031, LP-220100527, DP-220102121, and FT-220100318. 
We would like to thank anonymous reviewers for their constructive feedback that improved our paper.

{\small
\bibliographystyle{ieee_fullname}

}

\end{document}